\documentclass[10pt,twocolumn,letterpaper]{article}

\usepackage{cvpr}
\usepackage{times}
\usepackage{epsfig}
\usepackage{graphicx}
\usepackage{amsmath}
\usepackage{amssymb}

\usepackage{booktabs}

\usepackage[pagebackref=true,breaklinks=true,letterpaper=true,colorlinks,bookmarks=false]{hyperref}

\def\eg{\emph{e.g. }}
\def\ie{\emph{i.e. }}

\cvprfinalcopy 

\ifcvprfinal\fi
\begin{document}

\title{Task-driven Visual Saliency and Attention-based Visual Question Answering}

\author{Yuetan Lin \qquad Zhangyang Pang \qquad Donghui Wang\thanks{Corresponding author} \qquad Yueting Zhuang\\
College of Computer Science, Zhejiang University\\
Hangzhou, P. R. China\\
{\tt\small \{linyuetan,pzy,dhwang,yzhuang\}@zju.edu.cn}
}

\maketitle

\begin{abstract}
Visual question answering (VQA) has witnessed great progress since May, 2015
as a classic problem unifying visual and textual data into a system.
Many enlightening VQA works explore deep into the image and question encodings and fusing methods,
of which attention is the most effective and infusive mechanism.
Current attention based methods focus on adequate fusion of visual and textual features,
but lack the attention to where people focus to ask questions about the image.
Traditional attention based methods attach a single value to the feature at each spatial location,
which losses many useful information.
To remedy these problems, we propose a general method to perform saliency-like pre-selection on overlapped region features by the interrelation of bidirectional LSTM (BiLSTM),
and use a novel element-wise multiplication based attention method to capture more competent correlation information between visual and textual features.
We conduct experiments on the large-scale COCO-VQA dataset and analyze the effectiveness of our model demonstrated by strong empirical results.
\end{abstract}

\section{Introduction}
Visual question answering (VQA) comes as a classic task which combines visual and textual modal data into a unified system.
Taking an image and a natural language question about it as input,
a VQA system is supposed to output the corresponding natural language answer.
VQA problem requires image and text understanding, common sense and knowledge inference.
The solution of VQA problem will be a great progress in approaching the goal of Visual Turing Test,
and is also conducive to tasks such as multi-modal retrieval, image captioning and accessibility facilities.

\begin{figure}[!htp]
\centering
\includegraphics[width=0.47\textwidth]{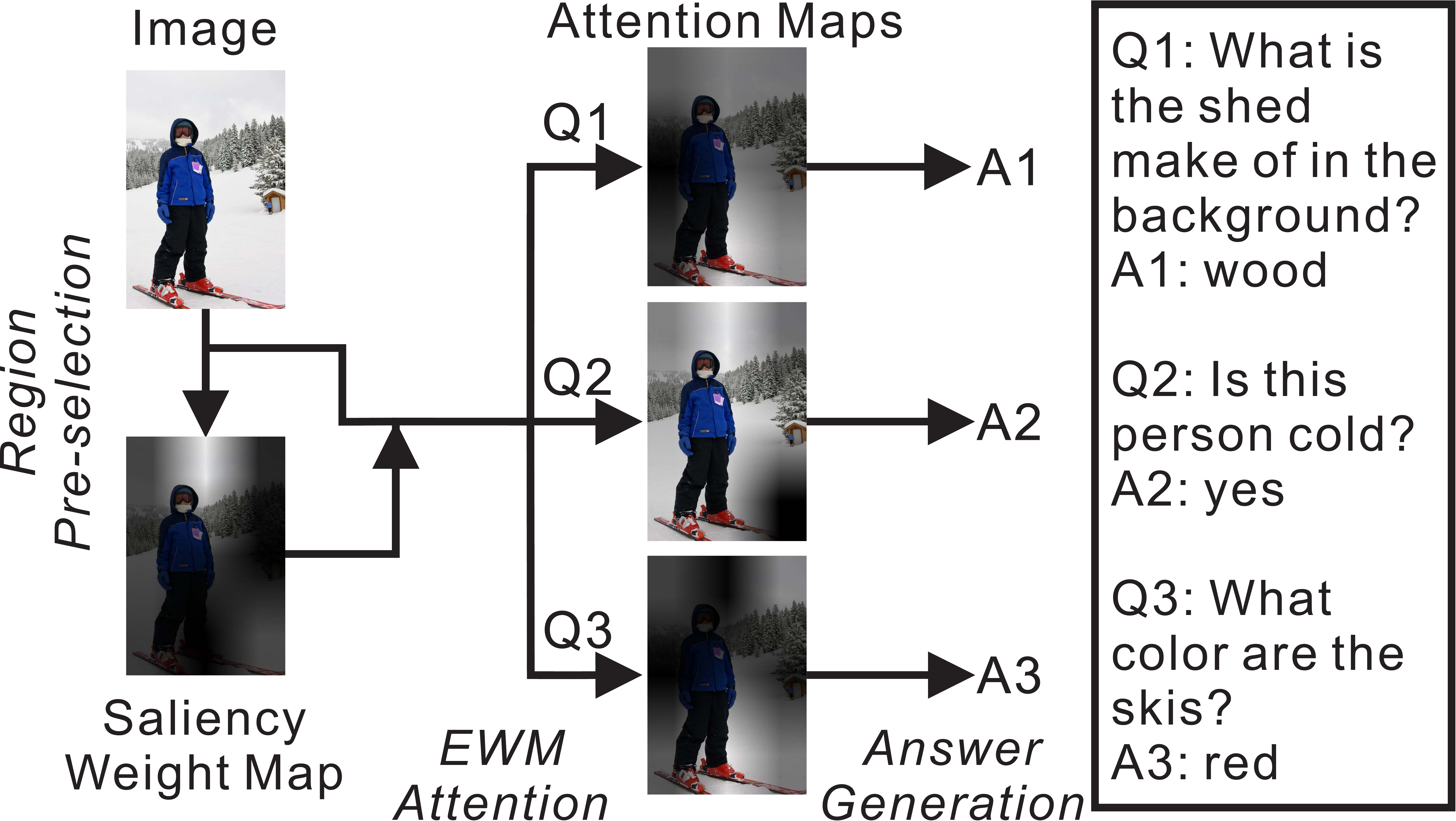}
\caption{(Best view in color and zoom in.)
The flow of our proposed VQA model.
Q and A represent question and answer related to the image.
EWM denotes element-wise multiplication operation.
}
\label{fig:demo}
\end{figure}

After the first attempt and introduction of VQA~\cite{malinowski2014multi},
more than thirty works on VQA have sprung up over the past one year from May, 2015.
Over ten VQA datasets and a big VQA challenge~\cite{antol2015vqa} have been proposed so far.
Four commonly used datasets
(\ie DAQUAR~\cite{malinowski2014multi}, COCO-QA~\cite{ren2015exploring}, COCO-VQA~\cite{antol2015vqa} and Visual7W~\cite{zhu2016visual7w}) feature different aspects.
The common practice to tackle VQA problem is to translate the words as word embeddings and encode the questions using bag-of-word (BoW)
or Long Short Term Memory (LSTM) network,
and encode the images using deep convolutional neural networks (CNN).
The following important step is to combine the image and question representations through some kind of fusing methods for answer generation,
such as concatenation~\cite{zhou2015simple,malinowski2015ask,gao2015you},
element-wise multiplication~\cite{antol2015vqa}, parameter prediction layer~\cite{noh2016image},
episode memory~\cite{xiong2016dynamic}, attention mechanism~\cite{shih2016look,kim2016multimodal,lu2016hierarchical}, etc.
Current VQA works focus on the fusion of these two features,
yet no one cares about ``where we focus'' to ask questions on the image.
It is a common practice to treat the VQA problem as either a generation, classification or a scoring task,
and classification gains more popularity due to its simplicity and easiness for comparison.

These works treat VQA as a discriminative model,
learning the conditional probability of answer given the image and question.
From the generative view,
we emulate the behavior that before people ask questions about the given image they first glance at it and find some interesting regions.
In terms of a single person, he has unique taste for choosing image regions that interest him.
For a large amount of people, there are statistical region-of-interest (RoI) distributions.
These region patterns are task-driven,
\eg the picture in Figure.~\ref{fig:demo},
for VQA task people may focus mostly on the beds, the chairs, the laptop and the notebook regions (namely the RoI patterns) as captured in the weighted image,
but for image captioning task they pay attention to more areas including the striped floor.
It is very valuable to intensify the interesting region features and suppress others,
and this image preprocessing step provides more accurate visual features to the follow-up steps and is missing in current VQA works.
By analogy with visual saliency which captures the standing out regions or objects of an image,
we propose a region pre-selection mechanism named task-driven visual saliency which attaches interesting regions (more possibly questioned on) with higher weights.
Taking advantage of the bidirectional LSTM (BiLSTM)~\cite{graves2005framewise}
that the output at an arbitrary time step has complete and sequential information about all time steps before and after it,
we compute the weight of interest for each region feature which is relative to all of them.
To the best of our knowledge, this is the first work that employs and analyzes BiLSTM in VQA models for task-driven saliency detection,
and this is the first contribution of our work.

As a simple and effective VQA baseline method,
\cite{zhou2015simple} shows that question feature always contributes more to predict the answer than image feature.
But image is as equally important as question for answer generation.
It is necessary to further explore finer-grained image features to achieve better VQA performance,
\eg attention mechanism \cite{xu2015show}.
Current attention based models generally use the correlation scores between question and image representations as weights to perform weighted sum of region features,
the resulting visual vector is concatenated to the question vector for final answer generation.
The recent ``multi-step'' attention models (\ie containing multiple attention layers) \cite{yang2016stacked,lu2016hierarchical}
dig deeper into the image understanding and
help achieve better VQA performance than the ``regular'' attention models.
However, the correlation score obtained by inner product between visual and textual features is essentially the sum of the correlation vector obtained by element-wise multiplication of the two features.
Besides, \cite{antol2015vqa} shows that element-wise multiplication of these features achieves more accurate results than concatenation of them in the baseline model.
Hence we propose to employ element-wise multiplication way in the attention mechanism,
the fused features are directly feed forward to a max pooling layer to get the final fused feature.
Together with the saliency-like region pre-selection operation,
this novel attention method effectively improves VQA performance and is the second contribution of this work.


The remainder of the paper is organized as follows.
We first briefly review saliency and the attention mechanism.
Then, we elaborate our proposed method.
We present experiments of some baseline models and compare with state-of-the-art models and visualize the pre-selection saliency and attention maps.
Finally we summarize our work.

\section{Related Work}
\subsection{Saliency Detection Modeling}
Saliency generally comes from contrasts between a pixel or an object and its surroundings,
describing how outstanding it is.
It could facilitate learning by focusing the most pertinent regions.
Saliency detection methods mimic the human attention in psychology,
including both bottom-up and top-down manners~\cite{van1997task}.
Typical saliency methods~\cite{itti1998model,liu2011learning} are pixel- or object-oriented,
which are not appropriate for VQA due to center bias and are difficulty in collecting large scale eye tracking data.

We think task-driven saliency on image features could be conductive to solving VQA problem.
What inspires us is that BiLSTM used in saliency detection has achieved good results on text and video tasks.
In sentiment classification tasks,
\cite{li2016visualizing} assigns saliency scores to words related to sentiment for visualizing and understanding the effects of BiLSTM in textual sentence.
While in video highlight detection,
\cite{yang2015unsupervised} uses a recurrent auto-encoder configured with BiLSTM cells and extracts video highlight segments effectively.
BiLSTM has demonstrated its effectiveness in saliency detection,
but to the best of our knowledge it has not been used in visual saliency for VQA task.


\subsection{Attention in VQA Models}
Visual attention mechanism has drawn great interest in VQA~\cite{yang2016stacked,zhu2016visual7w,shih2016look}
and gained performance improvement from traditional methods using holistic image features.
Attention mechanism is typically the weighted sum of the image region features at each spatial location,
where the weights describe the correlation and are implemented as the inner products of the question and image features.
It explores finer-grained visual features and mimics the behavior that people attend to different areas according to the questions.
Focusing on ``knowing where to look'' for multiple-choice VQA tasks,
\cite{shih2016look} uses 99 detected object regions plus a holistic image feature to make correlation with the question encoding,
and uses the correlation scores as weights to fuse the features.
\cite{yang2016stacked} uses the last pooling layer features ($512\times14\times14$) of VGG-19~\cite{simonyan2015very} as image region partitions,
and adopts two-layer attention to obtain more effective fused features for complex questions.
\cite{andreas2016neural} proposes an ingenious idea to use assembled network modules according to the parsed questions,
and achieves multi-step transforming attention by specific rules.

However, these attention methods use correlation score (\ie inner product between visual and textual feature) for each location,
which is the sum of the correlation vector representation (\ie element-wise multiplication between them).
Besides, the concatenation of image and question features is less accurate than the element-wise multiplication vector of them shown in the baseline model~\cite{antol2015vqa}.
Moreover, there are many answers derived from non-object and background regions, \eg questions about scenes,
hence it is not fit for the object detection based attention methods.

\begin{figure*}[!ht]
\centering
\includegraphics[width=0.9\textwidth]{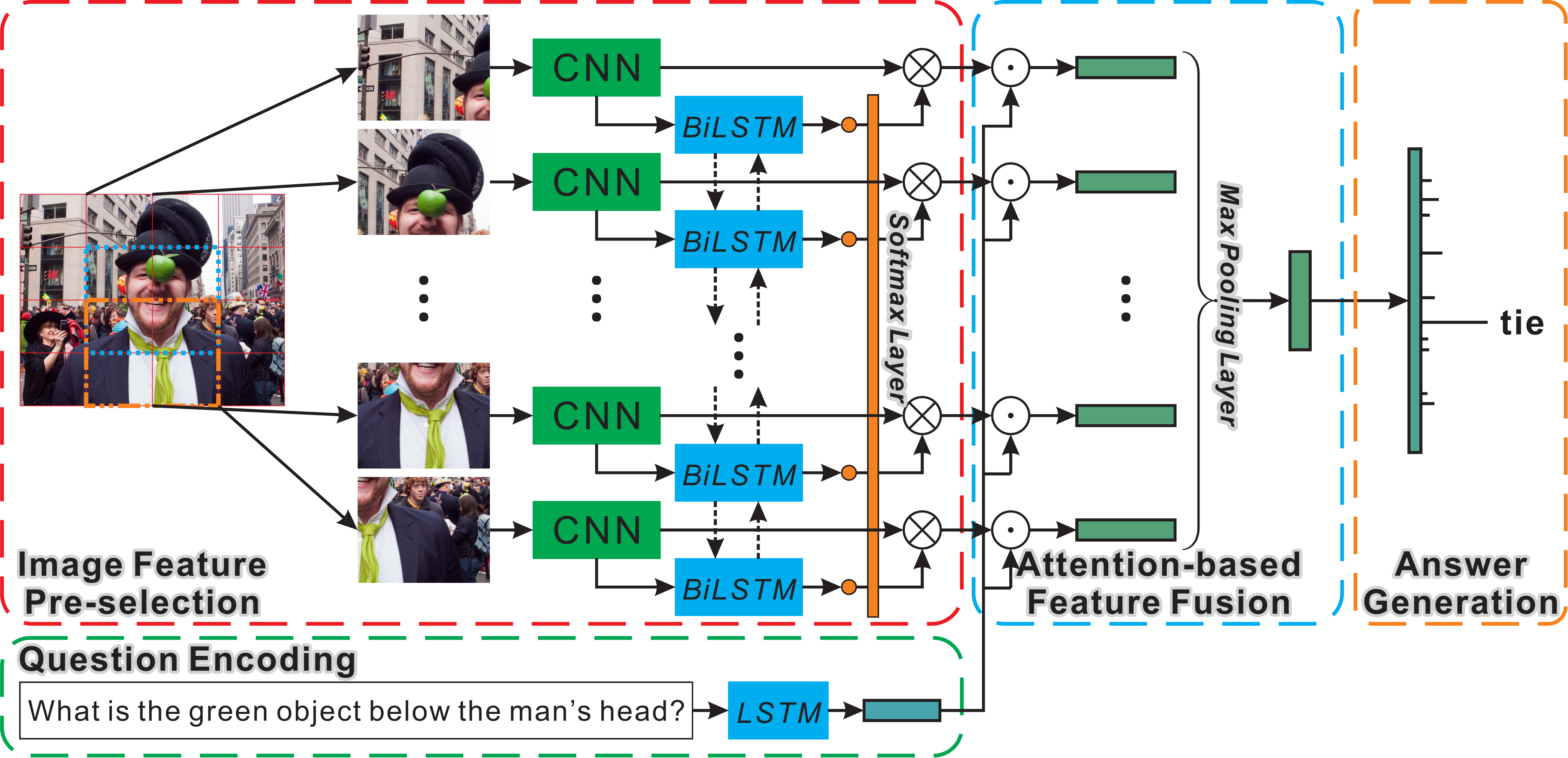}
\caption{(Best view in color and zoom in.)
Frameworks of our proposed VQA model.
Bars with frame lines denote vectors and small circles denote scalars.
$\bigotimes$ and $\bigodot$ represent multiplication of scalar and vector and element-wise multiplication of two vectors, respectively.
}
\label{fig:framework}
\end{figure*}

\section{Proposed Method}
Compared with image captioning which generates general descriptions about an image,
VQA focuses on specific image regions depending on the question.
On the one hand, these regions include non-object and background contents which are hard for object detection based VQA methods.
On the other hand, although people may ask questions at any areas of a given image,
there are always some region patterns that attract more questions.
On the whole, there are statistical region-of-interest (RoI) patterns which represent human-interested areas that are important for later VQA task.
We propose a saliency-like region pre-selection and attention-based VQA framework illustrated in Figure.~\ref{fig:framework}.
The VQA is regarded as a classification task, which is simple and easy to transform to a generating or scoring model.


\subsection{Model}
In this section, we elaborate our model consisting of four parts:
(a)~image feature pre-selection part which models the tendency where people focus to ask questions,
(b)~question encoding part which encodes the question words as a condensed semantic embedding,
(c)~attention-based feature fusion part performs second selection on image features and
(d)~answer generation part which gives the answer output.

\subsubsection{Image Feature Pre-selection}
As described above,
current object detection based VQA methods may not be qualified and the answers may not be derived from these specific object regions in images,
for example, when asked ``Where is the bird/cat?'', 
the answers ``fence/sink'' are not contained in ILSVRC~\cite{russakovsky2015imagenet} (200 categories) and Pascal VOC~\cite{everingham2010pascal} (20 categories) detection classes. 
Thus we use a more general pattern detector.

In addition,
from the generative perspective,
we pay attention to where people focus to ask questions.
General visual saliency provides analogous useful information of noticeable objects or areas which outstand the surroundings,
but it is not the only case for VQA task.
Current attention mechanism relates the question to the focusing location.
As more samples are available, we could yield the region patterns that attract more questions by statistics.
From the statistical behavior of large amounts of workers on Amazon Mechanical Turk (AMT) who have labeled the questions,
we model the region-of-interest patterns that could attract more questions.


We propose to perform saliency-like pre-selection operation to alleviate the problems and model the RoI patterns.
The image is first divided into $g\times g$ grids as illustrated in Figure.~\ref{fig:framework}.
Taking $m\times m$ grids as a region, with $s$ grids as the stride,
we obtain $n\times n$ regions, where $n=\left\lfloor\frac{g-m}{s}\right\rfloor+1$.
We then feed the regions to a pre-trained ResNet~\cite{he2016deep} deep convolutional neural network to produce $n\times n\times d_I$-dimensional region features,
where $d_I$ is the dimension of feature from the layer before the last fully-connected layer.

Since the neighboring overlapped regions share some visual contents,
the corresponding features are related but focusing on different semantic information.
We regard the sequence of regions as the result of eye movement when glancing at the image,
and these regions are selectively allocated different degrees of interest.
Specifically, the LSTM is a special kind of recurrent neural network (RNN),
capable of learning long-term dependencies via the memory cell and the update gates,
which endows itself with the ability to retain information of previous time-steps (\ie the previous region sequence in this case).
The update rules of the LSTM at time step $t$ are as follows:
\begin{align}
i_t&=\sigma(W^{(i)}x_t+U^{(i)}h_{t-1}+b^{(i)}),\label{eq:vqaF1}\\
f_t&=\sigma(W^{(f)}x_t+U^{(f)}h_{t-1}+b^{(f)}),\label{eq:vqaF2}\\
o_t&=\sigma(W^{(o)}x_t+U^{(o)}h_{t-1}+b^{(o)}),\label{eq:vqaF3}\\
u_t&=\tanh(W^{(u)}x_t+U^{(u)}h_{t-1}+b^{(u)}),\label{eq:vqaF4}\\
c_t&=u_t\odot i_t+c_{t-1}\odot f_t,\label{eq:vqaF5}\\
h_t&=o_t\odot \tanh(c_t),\label{eq:vqaF6}
\end{align}
where $i,f,o$ denote the input, forget and output gates,
$x,c,h$ are the input region feature, memory cell and hidden unit output,
and $W,U,b$ are the parameters to be trained.
We activate the gates by the sigmoid nonlinearity $\sigma(x)=1/(1+e^{-x})$
and the cell contents by the hyperbolic tangent $\tanh(x)=(e^x-e^{-x})/(e^x+e^{-x})$.
The gates control the information in the memory cell to be retained or forgotten through element-wise multiplication $\odot$.

Inspired by the information completeness and high performance of BiLSTM,
we encode the region features in two directions using BiLSTM and obtain a scalar output per region.
The output of the BiLSTM is the summation of the forward and backward LSTM outputs at this region location:
$h_t=h_t^{(f)}+h_{n-t+1}^{(b)}$, where $n$ is the number of regions, $h_t^{(f)},h_{n-t+1}^{(b)}$ are computed using Eq.~\ref{eq:vqaF6}.
Hence, the output at each location is influenced by the region features before and after it,
which embodies the correlation among these regions.
Note that, although the DMN+ work~\cite{xiong2016dynamic} uses similar bi-directional gated recurrent units (BiGRU) in the visual input module,
their purpose is to produce input facts which contain global information.
Besides, their BiGRU takes the features embedded to the textual space as inputs.
In contrast, the BiLSTM used in our model takes directly visual CNN features as input,
and the main purpose is to output weights for region feature selection.

The output values of the BiLSTM are normalized through a softmax layer,
and the resulting weights are multiplied by the region features.
We treat the weights as degree of interest which are trained by error back-propagation of the final class cross entropy losses,
and higher weights embody that the corresponding region patterns will attract more questions,
in other words, these region patterns may get higher attention values in the latter interaction with question embeddings in a statistical way.

\subsubsection{Question Encoding}
Question can be encoded using various kinds of natural language processing (NLP) methods,
such as BoW, LSTM, CNN \cite{ma2016learning,yang2016stacked}, gated recurrent units (GRU) \cite{cho2014properties}, skip-thought vectors \cite{kiros2015skip},
or it can be parsed by Stanford Parser \cite{klein2003accurate}, etc.
Since question BoW encodings already dominate the contribution to answer generation compared with the image features \cite{zhou2015simple},
we simply encode the question word as word2vec embedding, 
and use LSTM to encode the questions to match the pre-selected region features. 
To encode more abstract and higher-level information and achieve better performance,
a deeper LSTM \cite{antol2015vqa,karpathy2016visualizing} for question encoding is adopted in our model.

The question encoding LSTM in our model has $l$ hidden layers with $r$ hidden units per layer,
and the question representation is the last output and the cell units of the LSTM,
and the dimension is $d_Q=2\times l\times r$.
The resulting condensed feature vector encodes the semantic and syntactic information of the question.

\subsubsection{Attention-based Feature Fusion}
According to the statistic image-question-answer (IQA) training triples,
the image feature pre-selection has attached the regions with different prior weights,
generating more meaningful region features.
But different questions may focus on different aspects of the visual content.
It is necessary to use attention mechanism to second select regions by the question for more effective features.

We propose a novel attention method,
which takes the element-wise multiplication vector as correlation between image and question features at each spatial location.
Specifically, given the pre-selected region features and question embedding,
we map the visual and textual features into a common space of $d_C$ dimension and perform element-wise multiplication between them.
The $n\times n\times d_C$-dimensional fused features contain visual and textual information,
and higher responses indicate more correlative features.
In traditional attention models, the correlation score (scalar) achieved by inner product between the mapped visual and textual features per region,
is essentially the sum of elements in our fused feature.
This novel attention method has two noticeable advantages against traditional attention,
\ie information richer correlation vector versus correlation scalar,
more effective element-wise multiplication vector versus the concatenated vector of the visual and textual features.

Since higher responses in the fused features indicate more correlative visual and textual features,
and the question may only focus on one or two regions.
We choose to apply max pooling operation on the intermediate fused features to pick out the maximum responses.
The produced $d_C$-dimensional fused feature is then fed to the final answer generation part.
Compared to the sum/average operation in traditional attention models,
the max operation highlights the responses of the final fused feature from every spatial location.

\subsubsection{Answer Generation}
Taking the VQA problem as a classification task is simple to be implemented and evaluated,
and it is easy to be extended to generation or multiple choice tasks through a network surgery using the fused feature in the previous step.
We use a linear layer and a softmax layer to map from the fused feature to the answer candidates,
of which the entries are the top-1000 answers from the training data.

Considering multiple choice VQA problems,
\eg Visual7W~\cite{zhu2016visual7w} telling questions and COCO-VQA~\cite{antol2015vqa} multiple choice tasks,
our model is adaptive to be extended by concatenating the question and answer vectors before fusion with visual features
or by using bilinear model between the final fused feature and answer feature~\cite{shih2016look,jabri2016revisiting},
which is a possible future work.
Meanwhile, in view of generation VQA problem,
we can train an LSTM taking the fused feature as input to obtain answer word lists, phrases or sentences~\cite{malinowski2015ask,gao2015you}.

\subsection{Training}
Our framework is trained end-to-end using back-propagation,
while the feature extraction part using ResNet is kept fixed to speed up training and avoid the noisy gradients back-propagated from the LSTM as elaborated in~\cite{gao2015you}.
RMSprop algorithm is employed with low initial learning rate of 3e-4 which is proved important to prevent the softmax from spiking too early and prevent the visual features from dominating too early~\cite{shih2016look}.
Due to simplicity and proved similar performance as pre-trained word embedding parameters,
we initialize the parameters of the network with random numbers.
We randomly sample 500 IQA triples per iteration.

\section{Experiments}
In this section, we describe the implementation details and evaluate our model (SalAtt) on the large-scale COCO-VQA dataset.
Besides, we visualize and analyze the role of pre-selection and the novel attention method.

\subsection{Implementation Details}
In our experiment, the input images are first scaled to $448\times448\times3$ pixels before we apply $4\times4$ grids on them.
We obtain $3\times3$ regions by employing $2\times2$ grids (\ie $224\times224\times3$ pixels) as a region with stride $1$ grid.
Then we extract the $2048$-D feature per region from the layer before the last fully-connected layer of ResNet.
The dimension of word embedding is $200$,
and the weights of the embedding are initialized randomly from a uniform distribution on $[-0.08,0.08)$ due to similar performance as the pre-trained one.
The pre-selection BiLSTM for region features has $1$ layer and the size is $1$,
and the LSTM for question uses $2$ layers and $512$ hidden units per layer.
The common space of visual and textual features is $1024$-dimensional.
We use dropout~\cite{srivastava2014dropout} after all convolutional and linear layers.
The non-linear function is hyperbolic tangent.

The training procedure is early stopped when there is no accuracy increase in validation set for $5,000$ iterations where we evaluate every $1,000$ iterations.
It takes around 18 hours to train our model on a single NVIDIA Tesla K40 GPU for about $91,000$ iterations.
And for evaluation, each sample needs less than 0.5 millisecond.

\subsection{Datasets}
The COCO-VQA dataset~\cite{antol2015vqa} is the largest among the commonly used VQA datasets,
which contains two tasks (\ie multiple-choice task and open-ended task) on two image datasets (\ie real image MSCOCO dataset~\cite{lin2014microsoft} and abstract scene dataset).
We follow the common practice to evaluate models on two tasks on the real image dataset,
which includes 248,349 training questions, 121,512 validation questions and 244,302 testing questions.
There are many types of questions which require image and question understanding, commonsense knowledge, knowledge inference and even external knowledge.
The answers are roughly divided into 3 types, \ie ``yes/no'', ``number'' and ``other''.

To evaluate the results, each answer is compared with 10 human-labeled answers,
the accuracy is computed via this metric: $min(\frac{\#consistent\ human-labeled\ answers}{3},1)$,
\ie the accuracy is $100\%$ if the predicted answer is consistent with at least 3 human-labeled answers.
The COCO-VQA dataset provide human-labeled answers for the training and validation sets,
and the results of testing set can only be tested on the evaluation server.
The whole testing set is named test-standard and can be evaluated once per day and 5 times in total,
and a smaller development set named test-dev can be tested 10 times per day and 9999 times in total.
In short, the COCO-VQA dataset is large and hard enough for evaluating models and hence we choose to evaluate our model on it.

\subsection{Compared Models}
We compare our propose model (SalAtt) with some function-disabled models listed below,
to prove the effectiveness of the region pre-selection via BiLSTM and the novel attention method.

\begin{itemize}
\item \textbf{holistic:} The baseline model which maps the holistic image feature and LSTM-encoded question feature to a common space and perform element-wise multiplication between them.
\item \textbf{TraAtt:} The traditional attention model, implementation of WTL model~\cite{shih2016look} using the same $3\times3$ regions in SalAtt model.
\item \textbf{RegAtt:} The region attention model which employs our novel attention method, same as the SalAtt model but without region pre-selection.
\item \textbf{ConAtt:} The convolutional region pre-selection attention model which replaces the BiLSTM in SalAtt model with a weight-sharing linear mapping, implemented by a convolutional layer.
\end{itemize}

Besides, we also compare our SalAtt model with the popular baseline models \ie iBOWIMG~\cite{zhou2015simple}, VQA~\cite{antol2015vqa},
and the state-of-the-art attention-based models
\ie WTL~\cite{shih2016look}, NMN~\cite{andreas2016neural}, SAN~\cite{yang2016stacked}, AMA~\cite{wu2016ask}, FDA~\cite{ilievski2016focused}, D-NMN~\cite{andreas2016learning}, DMN+~\cite{xiong2016dynamic}
on two tasks of COCO-VQA.

\subsection{Results and Analysis}
\subsubsection{Effectiveness of Proposed Functions}

\begin{table}
\centering
\begin{tabular}{@{}cccccc@{}}
\toprule
Model & holistic & TraAtt & RegAtt & ConAtt & \textbf{SalAtt} \\
\midrule
VGG19  & 59.52 & 58.39 & 59.76 & 59.24 & \textbf{59.86} \\
ResNet & 60.30 & 59.07 & 60.45 & 60.25 & \textbf{60.62} \\
\bottomrule
\end{tabular}
\caption{Comparison of different models using two CNN features on COCO-VQA validation set, where SalAtt is our proposed model. (accuracy in \%)}
\label{tab:modelsres}
\end{table}

We train the function-disabled models on COCO-VQA training set and show the accuracies on validation set in Table.~\ref{tab:modelsres}.
From the columns, we can see that:
(1) holistic is better than TraAtt, proving the effectiveness of element-wise multiplication feature fusion compared with concatenation of features.
(2) RegAtt is better than holistic, indicating our novel attention method indeed enriches the visual features and improves the performance.
(3) SalAtt is better than RegAtt, demonstrating the strength of our region pre-selection mechanism.
(4) ConAtt is worse than SalAtt, showing that BiLSTM is important for the region pre-selection part.
From each row, we find the consistent improvement by the ResNet features,
showing the importance of good CNN features to VQA.

\subsubsection{Quantitative Results on Evaluation Server}

\begin{table*}[!ht]
\centering
\begin{tabular}{@{}cccccccccc@{}}
\toprule
        & \multicolumn{4}{c}{Open Ended}  && \multicolumn{4}{c}{Multiple Choice} \\ \cmidrule[.75pt]{2-5} \cmidrule[.75pt]{7-10}
        & All   & Yes/No & Number & Other && All   & Yes/No & Number & Other     \\ \midrule
iBOWIMG & 55.72 & 76.55 & 35.03 & 42.62 && 61.68 & 76.68 & 37.05 & 54.44 \\
VQA     & 57.75 & 80.50 & 36.77 & 43.08 && 62.70 & 80.52 & 38.22 & 53.01 \\
WTL     & -     & -     & -     & -     && 62.44 & 77.62 & 34.28 & 55.84 \\
NMN     & 58.60 & \textbf{81.20} & 38.00 & 44.00 && -     & -     & -     & -     \\
SAN     & 58.70 & 79.30 & 36.60 & 46.10 && -     & -     & -     & -     \\
AMA     & 59.17 & 81.01 & 38.42 & 45.23 && -     & -     & -     & -     \\
FDA     & 59.24 & 81.14 & 36.16 & 45.77 && \textbf{64.01} & \textbf{81.50} & \textbf{39.00} & 54.72 \\
D-NMN   & 59.40 & 81.10 & \textbf{38.60} & 45.50 && -     & -     & -     & -     \\
DMN+    & \textbf{60.30} & 80.50 & 36.80 & \textbf{48.30} && -     & -     & -     & -     \\ \midrule
Ours    & 58.97 & 79.39 & 36.29 & \textbf{46.67} && \textbf{63.69} & 79.42 & 38.12 & \textbf{56.02} \\ \bottomrule
\end{tabular}
\caption{Accuracies for the open-ended and multiple-choice tasks on VQA test-dev evaluated on the VQA Server.}
\label{tab:test-dev}
\end{table*}

\begin{table*}[!ht]
\centering
\begin{tabular}{@{}cccccccccc@{}}
\toprule
        & \multicolumn{4}{c}{Open Ended}  && \multicolumn{4}{c}{Multiple Choice} \\ \cmidrule[.75pt]{2-5} \cmidrule[.75pt]{7-10}
        & All   & Yes/No & Number & Other && All   & Yes/No & Number & Other     \\ \midrule
iBOWIMG & 55.89 & 76.76 & 34.98 & 42.62 && 61.97 & 76.86 & 37.30 & 54.60 \\
VQA     & 58.16 & 80.56 & 36.53 & 43.73 && 63.09 & 80.59 & 37.70 & 53.64 \\
NMN     & 58.70 & -     & -     & -     && -     & -     & -     & -     \\
SAN     & 58.90 & -     & -     & -     && -     & -     & -     & -     \\
D-NMN   & 59.44 & 80.98 & \textbf{37.48} & 45.81 && -     & -     & -     & -     \\
AMA     & 59.44 & 81.07 & 37.12 & 45.83 && -     & -     & -     & -     \\
WTL     & -     & -     & -     & -     && 63.53 & 78.08 & 34.26 & \textbf{57.43} \\
FDA     & 59.54 & \textbf{81.34} & 35.67 & 46.10 && \textbf{64.18} & \textbf{81.25} & \textbf{38.30} & 55.20 \\
DMN+    & \textbf{60.36} & 80.43 & 36.82 & \textbf{48.33} && -     & -     & -     & -     \\ \midrule
Ours    & 59.26 & 79.42 & 36.35 & \textbf{47.00} && \textbf{63.99} & 79.45 & \textbf{37.93} & \textbf{56.42} \\ \bottomrule
\end{tabular}
\caption{Results for both tasks on VQA test-standard.}
\label{tab:test-std}
\end{table*}

We summarize the accuracies on test-dev in Table.~\ref{tab:test-dev} and the test-standard results in Table.~\ref{tab:test-std}.
Our results are comparative or higher than the attention based methods,
especially on multiple-choice tasks.
The results on answer type ``other'', which includes object and scene type questions,
demonstrate the competence of our model in RoI detection.

Note that,
we only apply the proposed region pre-selection mechanism to the basic VQA model~\cite{antol2015vqa},
it can be embedded into any other attention-based models to improve their performance.
Due to computation and training time,
we use only $3\times3$ regions compared with other attention-based methods (\eg $100$ or $14\times14$ region features).
Through observation, we find that many small objects could not be split by the $3\times3$ regions,
which is adverse to the counting questions and could be further improved and is a possible future work.

\subsubsection{Visualization of Intermediate Results}

\begin{figure*}[!ht]
\centering
\includegraphics[width=0.85\textwidth]{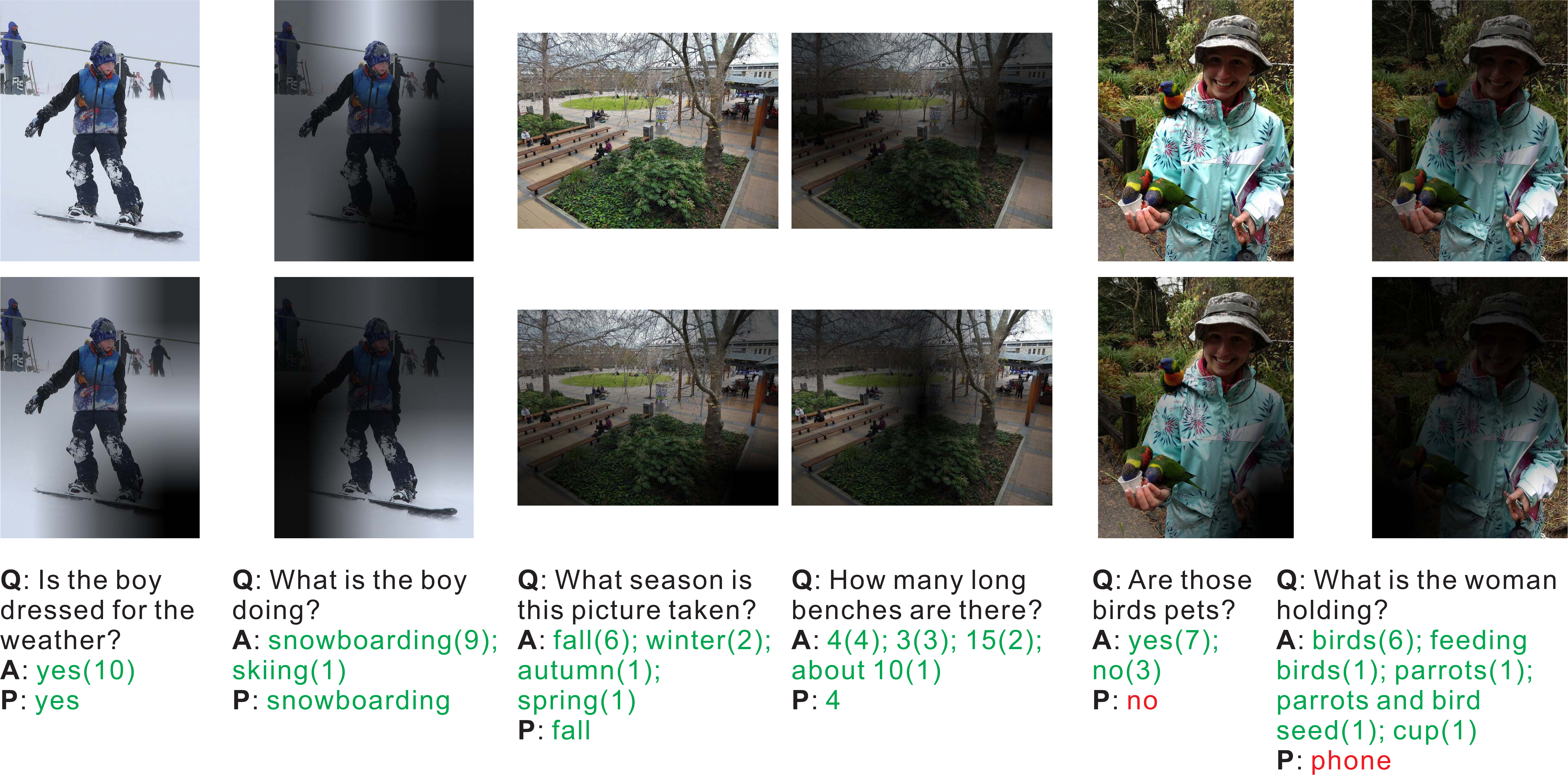}
\caption{(Best view in color and zoom in.)
Samples generated by our model.
The weights are normalized for visualization enhancement,
\ie the weight in the dark region may not necessarily be 0.
Q, A, P represent question, ground truth answer, predicted answer, respectively.
Green denotes true prediction and red false.
}
\label{fig:results}
\end{figure*}

We illustrate three groups of samples produced by our model in Figure.~\ref{fig:results}.
Each group contains four figures,
from left to right and from top to bottom,
they are respectively the original image, pre-selection weights on the image,
and two attention maps for different questions with the corresponding questions (Q), ground truth answers (A) and the predicted answers (P) shown below them.
And the number in the parentheses means the amount for this human-labeled answer entry.
The weights are normalized to have minimum 0 and maximum 1 for visualization enhancement,
\ie the weight in the dark region may not necessarily be 0.

Take the first sample for example,
the pre-selection operation gives high weight to the boy's head region which may be interesting to people and attract more questions (\eg questions containing the word ``boy'').
For the question ``Is the boy dressed for the weather?'',
the attention map focuses on the boy, his clothes and the surrounding regions to get a positive answer.
While for question ``What is the boy doing?'', it attends the boy and the snowboard, thus giving answer ``snowboarding''.
The third sample gives inaccurate but explainable answers,
\ie the birds may live in the park/zoo and come for food provided by the tourist so it may not be classified into pets,
and the left hand of the woman holds indeed a phone while the human-labeled answers focus on the right hand.

\section{Conclusion}
In this work, we propose a general VQA solution which integrates region pre-selection and a novel attention method to capture generic class region and richer fused feature representation.
These two procedures are independent, meanwhile they both contribute to better VQA performance.
Although the model is simple, it achieves comparative or higher empirical results than state-of-the-art models.

Possible future works include adopting finer-grained grids which capture more precise regions,
employing stacked attention layers for multi-step reasoning and more accurate answer location,
and applying the general pre-selection method to other attention-based VQA models.
The pre-selection mechanism is valuable and applicable to similar task, such as image captioning.


{\small
\bibliographystyle{ieee}
\bibliography{cvpr17}
}

\end{document}